\documentclass[pdflatex,sn-mathphys-num]{sn-jnl}

\usepackage{chngcntr}

\counterwithout{figure}{section}
\counterwithout{table}{section}

\renewcommand{\thefigure}{\arabic{figure}}
\renewcommand{\thetable}{\arabic{table}}
\usepackage{wrapfig}
\usepackage{graphicx}
\usepackage{caption}
\usepackage{multirow}
\usepackage{amsmath,amssymb,amsfonts}
\usepackage{amsthm}
\usepackage{mathrsfs}
\usepackage[title]{appendix}
\usepackage{xcolor}
\usepackage{textcomp}
\usepackage{manyfoot}
\usepackage{booktabs}
\usepackage{algorithm}
\usepackage{algorithmicx}
\usepackage{algpseudocode}
\usepackage{listings}

\theoremstyle{thmstyleone}%
\theoremstyle{thmstyletwo}%

\theoremstyle{thmstylethree}%

\usepackage{placeins}
\begin{document}

\title[Article Title]{Individual and Combined Effects of Synthetic ESL-Style and Typographical Variation on LLM Performance}


\author*[1]{\fnm{Serena} \sur{Liu}}\email{serenaliu@college.harvard.edu}
\author[1]{\fnm{Yutong} \sur{Yang}}
\author[1]{\fnm{Xinru} \sur{Zhu}}
\author[1]{\fnm{Prisha} \sur{Sheth}}
\equalcont{These authors contributed equally to this work.}
\author[1]{\fnm{Weixuan} \sur{Dong}}
\equalcont{These authors contributed equally to this work.}
\author[1]{\fnm{Mingjiao} \sur{Diao}}
\equalcont{These authors contributed equally to this work.}
\author[1]{\fnm{Nikhil} \sur{Banga}}
\author[1]{\fnm{Oscar} \sur{Melendez}}
\author[1]{\fnm{Leyang} \sur{Zeng}}
\author[1]{\fnm{Arnav} \sur{Sharma}}
\author[1]{\fnm{Minda} \sur{Zhao}}
\author[1]{\fnm{Marina} \sur{Lin}}
\author[1]{\fnm{Yangdong} \sur{Zhu}}
\author[1]{\fnm{Mengyu} \sur{Wang}}

\affil[1]{\orgname{Harvard University},
  \orgaddress{\city{Cambridge}, \state{Massachusetts},
  \postcode{02138}, \country{USA}}}

\abstract{Large language model (LLM) benchmarks often use clean benchmark English prompts. In practice, prompts may vary in grammar, word order, word choice, and spelling, and several forms of variation may occur together. Previous studies have usually examined English-as-a-second-language (ESL)-style variation and typographical errors separately. We ask whether robustness to either transformation describes model behavior when the two occur together. We evaluate four instruction-tuned models on six closed- and open-ended benchmarks under four matched conditions: clean benchmark English, synthetic first-language-informed ESL-style variation, typographical perturbation, and their combination. We also assess whether sampled transformations preserve the original task and compare automated evaluations with human judgments. The combined condition generally produces larger score decreases than either isolated condition. Its relation to the sum of the isolated effects varies across tasks and models. Closed-ended benchmarks are typically close to additive or subadditive, whereas open-ended benchmarks show greater variation across models and evaluation protocols. These findings show that robustness scores should be interpreted in the context of the inputs and evaluation procedures that produced them. Claims about model reliability should be limited to the conditions tested. Studies using synthetic transformations or automated scoring should show that the task is preserved and that the scoring procedure supports the intended interpretation.}





\maketitle

\section{Introduction}\label{sec1}

\subsection{Evidential Responsibility in LLM Evaluation}

\begin{figure}[!htbp]
    \centering
    \includegraphics[width=\linewidth]{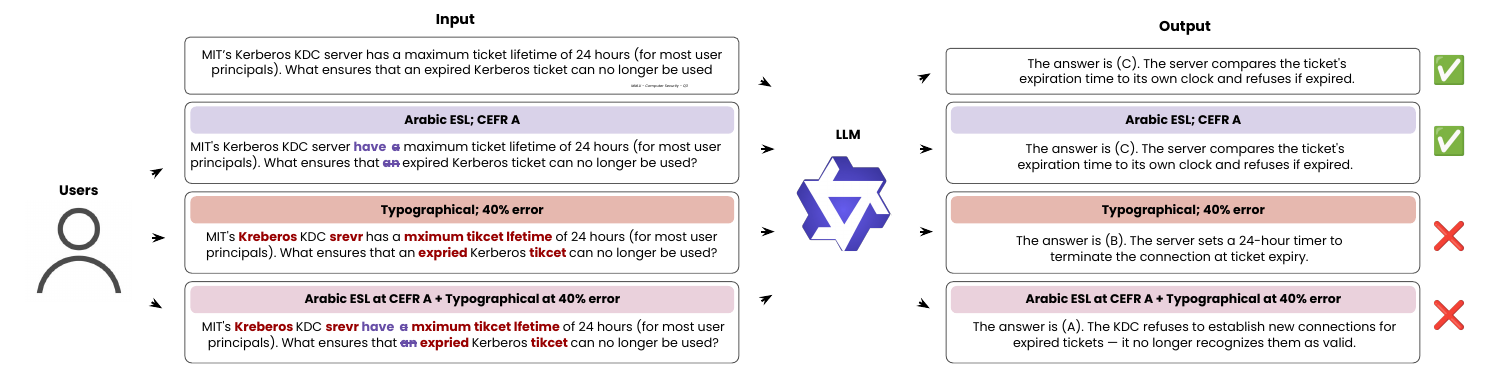}
    \caption{Sample LLM outputs on an MMLU question under four input conditions. ESL-style rewriting introduces controlled grammatical and lexical variation, while typographical perturbation alters surface forms. The combined condition illustrates how multiple forms of language variation may affect the same model response.}
    \label{fig:intro}
\end{figure}

Large language models (LLMs) are deployed globally as chatbots, translation tools, educational applications, and professional agents \citep{Brown+2020, Ouyang+2022, OpenAI2023}. Benchmark evaluations provide a controlled and reproducible way to compare model performance. At the same time, the inputs used in such evaluations cover only a limited range of linguistic conditions. English-language inputs may differ in syntax, article use, word order, lexical choice, spelling, and other surface characteristics, and more than one form of variation may occur within the same input.

Benchmark results are often used to support claims about model reliability, model selection, and deployment readiness, and they may also inform fairness-relevant evaluations \citep{mitchell2019model,mokander2024auditing}. The scope of those claims depends, in part, on the conditions included in the evaluation and on the procedures used to generate and score model outputs \citep{grote2024fairness,jacobs2021measurement}. A model that appears robust to one form of input variation may not behave in the same way when multiple forms of variation occur together. This leads to the central question of this study: does robustness observed under isolated input transformations characterize model behavior when those transformations are combined?

We examine this question using two synthetic input conditions:  ESL variation and typographical perturbation. The former introduces grammatical, lexical, and structural variation, whereas the latter alters the surface form of the input. These are experimental transformations, not representations of particular user populations or of the everyday distribution of typing errors. Accordingly, the study examines behavior under specified synthetic conditions and does not draw conclusions about naturally occurring ESL inputs or particular language communities.

Prior work has often examined linguistic variation and typographical perturbations separately, leaving their joint effects less well characterized. Separate analyses can show how performance changes under each transformation, but they do not establish how those changes relate to performance when both transformations are present. We therefore distinguish two questions: whether the combined condition causes more degradation than either isolated transformation, and whether the combined degradation is smaller than, close to, or larger than the sum of the isolated degradations. The latter relationship may differ across models, tasks, and evaluation procedures.  Figure~\ref{fig:intro} illustrates the four input conditions using an MMLU example.

To study these comparisons, we generate synthetic ESL-style transformations using Trans-EnV~\citep{Lee+2025} and typographical perturbations using MulTypo~\citep{Liu+2025}. For each source prompt, we construct four matched conditions: a clean Standard American English baseline, ESL-only variation, typo-only perturbation, and their combination. The underlying benchmark instance and evaluation protocol are held fixed across conditions, allowing direct comparison between isolated and combined performance changes. We evaluate four instruction-tuned language models across six benchmarks spanning closed-ended and open-ended tasks.

\begin{figure}[htbp]
    \centering
    \includegraphics[width=0.75\linewidth]{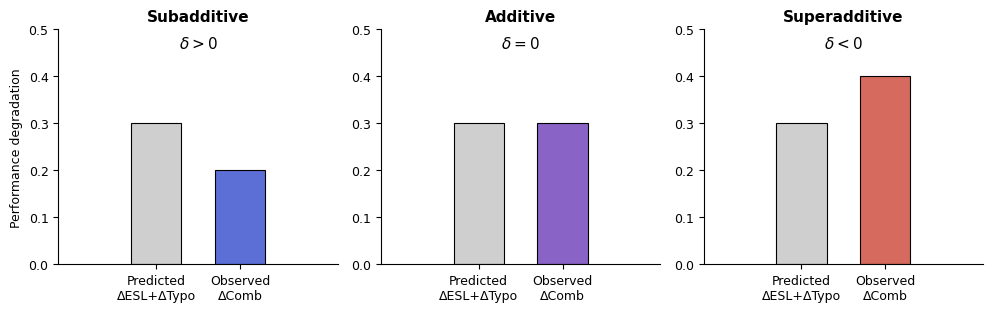}
    \caption{Illustration of interaction patterns under combined ESL-style and typographical perturbations. Gray bars denote the degradation predicted under an additive comparison, $\Delta_{\mathrm{ESL}}+\Delta_{\mathrm{Typo}}$, while colored bars denote the observed combined degradation, $\Delta_{\mathrm{Comb}}$.}
    \label{fig:interaction_framework}
\end{figure}

We summarize the second comparison using the interaction score $\delta=(\Delta_{\mathrm{ESL}}+\Delta_{\mathrm{Typo}})-\Delta_{\mathrm{Comb}}$, where each $\Delta$ denotes the score decrease relative to the clean baseline. Values near zero indicate approximately additive behavior; positive values indicate subadditive degradation, whereas negative values indicate superadditive degradation. Because benchmark scores are bounded, $\delta$ is interpreted descriptively rather than as evidence of a specific causal or cognitive mechanism. As shown in Fig.~\ref{fig:interaction_framework}, the observed combined degradation may fall below, match, or exceed the additive expectation.

The transformed inputs also introduce a measurement question. Synthetic perturbations do not necessarily preserve every aspect of the original task, particularly at higher levels of typographical corruption. In addition, some open-ended benchmarks rely on automated scoring procedures. We therefore assess task preservation in sampled ESL-style and typographical transformations and compare automated evaluation with independent human judgments for MT-Bench and AlpacaFarm. These analyses help interpret observed score differences in light of task recoverability and the agreement between automated and human evaluation.

The matched comparisons show that the combined condition generally produces larger score decreases than either isolated transformation, but the interaction pattern is not uniform. Closed-ended benchmarks are primarily additive or subadditive, whereas open-ended benchmarks show greater heterogeneity across models and evaluation protocols. These results indicate that performance under an isolated transformation does not always characterize model behavior when both transformations are present.

Benchmark results are evidence about model behavior under specified input conditions and evaluation procedures; by themselves, they do not establish reliability beyond those conditions. We therefore argue that evaluators have a responsibility to report the conditions covered by a test, whether transformed inputs preserve the task, and how automated scoring has been validated, particularly when benchmark results are used to support broader claims about model selection or deployment. Our study examines these reporting obligations under controlled synthetic conditions. Results from isolated ESL-style and typographical transformations do not consistently describe performance when the two are combined, while the validation analyses show why input coverage, task preservation, and score interpretation require distinct evidence.
\subsection{Contributions}

\noindent\textbf{(i) Matched evaluation of combined variation.}
We present a matched four-condition design. Each source input is evaluated under clean benchmark English, a synthetic ESL-style variant informed by first-language transfer, a typographically perturbed variant, and a variant that combines the two transformations. The same underlying benchmark item and evaluation procedure are used across all four conditions. This makes the isolated and combined conditions directly comparable.

\noindent\textbf{(ii) Descriptive analysis of combined variation.}
We use a descriptive interaction measure to compare the score decrease under the combined condition with the sum of the decreases under the two isolated conditions. The analysis covers four instruction-tuned models and six closed- and open-ended benchmarks. The combined condition does not follow a single pattern across tasks and models. Closed-ended benchmarks are generally close to additive or subadditive. Open-ended benchmarks show greater variation across models and scoring procedures. Results from clean inputs or single transformations therefore do not always characterize performance under the combined conditions studied here.

\noindent\textbf{(iii) Assessment of task preservation and scoring.}
We use human annotations to assess task preservation in sampled synthetic ESL-style and typographically perturbed inputs. For MT-Bench and AlpacaFarm, we conduct condition stratified audits that compare automated LLM-judge evaluations with human judgments. Together, these analyses clarify what the reported robustness scores can support and where their interpretation should be qualified.

These contributions inform a specific question about evidential
responsibility: what should evaluators disclose when results from isolated input transformations are used to support claims covering their combination? Model reporting and measurement research already emphasize the importance of documenting  evaluation conditions and examining the assumptions behind measured scores
\citep{mitchell2019model,jacobs2021measurement}.
We apply these concerns to combined-input evaluation, distinguishing
a gap in tested conditions from uncertainty about task preservation
and uncertainty about scoring. This distinction clarifies why
additional benchmark coverage alone is insufficient to support a
broader interpretation of robustness.

\section{Methods}

\subsection{Experimental design}

We examine model performance under controlled synthetic ESL-style variation and typographical perturbations, applied separately and in combination. We use these transformations as controlled diagnostic conditions to examine the coverage of benchmark evaluations and the interpretation of the scores they produce. They are not intended to reproduce the full distribution of naturally occurring ESL writing or to represent real language communities.

For each source prompt, we construct four matched conditions: clean benchmark English, synthetic ESL-style variation only, typographical perturbation only, and combined ESL-style plus typographical variation. The clean benchmark-English condition serves as the baseline. The ESL-style transformation is applied to the clean prompt, while typographical corruption is applied either directly to the clean prompt in the typo-only condition or after the ESL-style transformation in the combined condition.

Within each benchmark, the source task, labels, answer options or reference answers where applicable, decoding settings, scoring procedures, and evaluation protocol are held fixed across conditions. This matched design allows direct comparison of isolated and combined performance changes while keeping the underlying benchmark instance constant. Because transformed prompts may not always preserve the original task, particularly at higher typo rates, we interpret observed differences as associations with the specified transformation conditions, rather than as causal effects of surface-form sensitivity or evidence of a particular cognitive mechanism.

Alongside the main experiments, we conducted five validation studies, summarized in Table 1. Three assessed task preservation in sampled ESL-only, typo-only, and combined prompts; the typo-only study also assessed perceived typo realism. Two audits compared automated evaluations with human judgments on MT-Bench and AlpacaFarm.

\subsection{ESL transformation}

We use Trans-EnV \citep{Lee+2025} to generate controlled synthetic ESL-style variants from clean benchmark-English prompts. The framework applies rule-based rewrites parameterized by CEFR level and a transformation setting. The transformation categories used in this study include simplified syntax, altered word order, article omission, subject--verb agreement changes, and changes in lexical choice. These rule categories define the experimental manipulation and are not intended to provide a complete account of how people with a particular first language use English.

We treat the transformed prompts as controlled input perturbations rather than as direct samples of naturally occurring ESL writing or as representations of homogeneous user groups. The transformation labels identify transformation settings implemented by Trans-EnV. They do not identify the language background of a particular user or imply that the generated prompts are representative of a specific language community. The purpose of these transformations is to examine model behavior under specified grammatical, lexical, and structural variation.

We use the CEFR-A setting as a fixed and relatively demanding transformation level within the framework. The experiments include eight transformation settings: Arabic, French, German, Japanese, Mandarin, Portuguese, Russian, and Spanish. These settings were selected because they are available in Trans-EnV and provide variation in the rule configurations used to generate the transformed prompts.

The transformation procedure is designed to retain the meaning and task intent of the source prompt, but task preservation is assessed empirically rather than assumed. We conduct a separate human-validation study covering 360 transformed instances across the eight transformation settings and six benchmarks. Human ratings assess meaning preservation, key-information preservation, and readability. Core-content preservation is derived from the final meaning- and key-information-preservation labels, and the resulting preservation measures are summarized using benchmark-balanced macro-averages.
\subsection{Typographical perturbation} We use MulTypo~\citep{Liu+2025} to add controlled, character-level typographical perturbations to clean Standard American English (SAE) benchmark prompts and their matched ESL-style counterparts. It introduces four types of character-level edits: insertion, deletion, substitution, and transposition. Substitutions are constrained by keyboard layout to approximate common typing errors. These inputs are synthetic perturbations for controlled evaluation and are not intended to represent naturally occurring user writing. For each source prompt, we generate three perturbation levels with the MulTypo corruption parameter $r \in {0.1, 0.4, 0.7}$, corresponding to low, moderate, and severe perturbation. Larger values of $r$ produce more extensive editing. The same procedure is applied to the clean SAE prompts and their matched ESL-style counterparts: in the typo-only condition, perturbations are added to the clean prompt, whereas in the combined condition, they are added after the ESL-style transformation. We treat $r=0.7$ as a severe synthetic stress-test condition rather than a typical level of everyday typing noise. At higher levels, the perturbations may affect not only spelling but also readability and the recoverability of task-critical information. We therefore interpret model performance as a response to the complete generated input condition. Meaning preservation, key-information preservation, readability, and perceived typo realism are assessed in the human-validation study described below.

\subsection{Human validation of synthetic transformations}
\label{sec:human_validation}

To assess task preservation in the generated inputs, we conduct prompt-level validation for the ESL-only, typo-only, and combined conditions. Annotators compare each transformed prompt with its clean source prompt and assess meaning preservation, key-information preservation, and readability. The typo-only validation also assesses perceived typo realism. Because these inputs are synthetic, the analyses characterize the generated stimuli rather than naturally occurring ESL writing or typing behavior.

\paragraph{Sampling design.} We use a benchmark-stratified matched sample of 45 source prompts from six benchmarks. We randomly sample 10 prompts from each closed-ended benchmark (MMLU, GSM8K, and HellaSwag) and 5 prompts from each open-ended benchmark (MT-Bench, IFEval, and AlpacaFarm). We used the same benchmark-stratified set of 45 source prompts in both validation studies. Prompts were sampled without replacement using a fixed random seed of 42.

\paragraph{ESL-style validation.} For each selected source prompt, we included all eight ESL-style transformation settings, yielding 360 transformed instances across the six benchmarks. Human annotators assess meaning preservation, key-information preservation, and readability. Core-content preservation was not rated as a separate criterion. Instead, it was computed from the final adjudicated labels for meaning and key-information preservation. Preservation rates are summarized using benchmark-balanced macro-averages, with each benchmark contributing equally to the aggregate estimate.

\paragraph{Typographical validation.} For each of the same 45 source prompts, we retrieved one pre-generated typographical variant at each of three corruption levels, \(r \in \{0.1,0.4,0.7\}\), from a typo dataset created by our team. This yielded 135 variants in total. The $r=0.7$ condition is treated as a severe synthetic stress-test condition rather than as a typical level of everyday typing noise.

\paragraph{Combined-condition validation.}
To assess whether the original task remained recoverable after
both transformations, we selected two source prompts from each
of the six benchmarks. For each source prompt, we examined the
combined input under all eight ESL-style settings and three typo
rates, yielding 288 prompts. Annotators compared each combined
prompt with its clean source prompt and rated meaning preservation,
key-information preservation, and readability using the criteria
described above. Core-content preservation was calculated from
the final labels: a prompt counted as preserved only when both
meaning and key information were marked Yes.

\paragraph{Annotation criteria and procedure.} Meaning preservation indicates whether the transformed prompt retains the overall semantic content and task intent of the source prompt. Key-information preservation indicates whether the information required to interpret and complete the original task is retained. Both criteria are recorded using categorical Yes/No labels. For the preservation summaries, Yes is coded as preserved and No as not preserved. Core-content preservation is defined as:
\[
\mathrm{CorePreserved}_i =
\mathbb{1}\left(
\mathrm{Meaning}_i=\mathrm{Yes}
\land
\mathrm{KeyInfo}_i=\mathrm{Yes}
\right).
\]

Readability was rated on a three-point ordinal scale: 1 = difficult to understand, 2 = understandable with effort, and 3 = easy to understand. For typographical validation, typo realism was also rated on the same three-point scale, with higher values indicating errors that more closely resembled ordinary human typing mistakes.

Each transformed instance is independently rated by two proficient English-speaking members of the research team. Before rating, the raters receive the same instructions, criterion definitions, and examples. When the two raters agree, their shared rating is retained as the final label. Disagreements are reviewed by a third member of the research team, whose decision determines the final label used in the analysis.  For each transformation setting and typo severity level, preservation rates are first calculated within each benchmark and then averaged equally across the six benchmarks to obtain benchmark-balanced macro-averages.
\subsection{Models}

We evaluate four instruction-tuned large language models: GPT-5.2, DeepSeek-V3.2, Qwen2.5-7B-Instruct, and LLaMA-3-8B-Instruct. The model set includes both API-based and open-weight models, covering different model families and deployment settings.

For each model--benchmark pair, decoding settings, prompt format, and scoring procedures are kept fixed across the matched input conditions. The same configuration is used at each applicable typo rate. This keeps the evaluation setup consistent and makes performance comparisons across input conditions easier to interpret.

\subsection{Benchmarks}

We evaluate six benchmarks, including three closed-ended benchmarks and three open-ended benchmarks. The closed-ended benchmarks are MMLU~\citep{Hendrycks+2021}, GSM8K~\citep{Cobbe+2021}, and HellaSwag~\citep{Zellers+2019}. MMLU measures broad knowledge across multiple domains, GSM8K tests multi-step reasoning on grade-school mathematics problems, and HellaSwag tests commonsense reasoning through a multiple-choice continuation task.

The open-ended benchmarks are MT-Bench~\citep{Zheng+2023}, IFEval~\citep{Zhou+2023IFEval}, and AlpacaFarm~\citep{Dubois+2023}. MT-Bench evaluates multi-turn conversational response quality using LLM-based judgments. IFEval measures instruction following through verifiable output constraints, and AlpacaFarm evaluates response preferences through pairwise comparisons.

For closed-ended benchmarks, ESL-style and typographical transformations are applied to input questions, while answer options and gold labels remain unchanged. For open-ended benchmarks, transformations are applied to prompts or dialogue turns, and responses are evaluated using benchmark-specific procedures.

\subsection{Response Generation}

We generated responses using two inference setups. \texttt{Qwen/Qwen2.5-7B-Instruct} and \texttt{meta-llama/Meta-Llama-3-8B-Instruct} were run locally with Hugging Face Transformers. \texttt{deepseek-ai/DeepSeek-V3.2} was accessed through DeepInfra’s OpenAI-compatible API, while GPT-5.2 was accessed through the OpenAI API.

For closed-ended benchmarks, we used greedy decoding, with sampling disabled and a maximum of 100 new tokens per response. Outputs that could not be parsed into the required answer format were treated as incorrect, and no corrective retries were performed.

For MT-Bench and IFEval, the temperature was set to 0.5 by default and reduced to 0.15 for reasoning and mathematics prompts. The maximum completion length was 256 tokens, increased to 500 tokens for coding prompts. For AlpacaFarm, we used a temperature of 0.7 and a maximum completion length of 512 tokens.

 We allowed up to two additional attempts only when a response was incomplete or failed automatically checked formatting requirements. For MT-Bench, the full conversation history, including the model's first-turn response, was retained when generating the second-turn response.
\subsection{Evaluation protocols}

For each model--benchmark pair, we apply the same evaluation procedure across the four matched input conditions: clean benchmark English, ESL-only, typo-only, and ESL-plus-typo. Decoding and scoring settings are held fixed across conditions and, where applicable, typo rates. Scores are analyzed within the evaluation procedure associated with each benchmark.

For the closed-ended benchmarks, performance is measured by accuracy. For MMLU and HellaSwag, the predicted answer choice is compared with the corresponding gold label. For GSM8K, the model's final numerical answer is compared with the gold answer using the benchmark's answer-matching procedure.

For the open-ended benchmarks, we adapt each benchmark’s evaluation procedure to our experimental setting. In MT-Bench, an LLM judge evaluates each response turn separately, assigning an overall score and scores for helpfulness, correctness, depth, and clarity on a 0–5 scale. We use the overall score for the main analysis and report dimension-level scores where relevant. For IFEval, we use locally implemented deterministic checks and count a response as successful only if it satisfies all associated constraints. For AlpacaFarm, an LLM judge compares each model response with a fixed reference response for the same source item. The two responses are presented in random order. We map the judge’s preference back to the evaluated model response and assign a score of 1 for a win, 0 for a loss, and 0.5 for a tie.

\paragraph{Human validation of automated evaluation.} Because MT-Bench and AlpacaFarm rely on automated judgments, we separately assess their correspondence with human evaluation.

For MT-Bench, we audited 96 response turns in a condition stratified sample, with 24 observations from each of the Clean, ESL-only, Typo-only, and Combined conditions. Two members of the research team independently rated each response on helpfulness, correctness, depth, and clarity using the same 0--5 scale. For each dimension, ratings differing by fewer than two points were averaged. Ratings differing by two or more points were adjudicated by a third member of the research team (R3). The human composite score was calculated as the mean of the four resulting dimension scores.
The audit was not organized as 24 complete four-condition matched sets. Condition-specific judge--human comparisons are therefore descriptive and are not estimates from paired comparisons.

For AlpacaFarm, we audit 96 pairwise preference judgments involving Qwen2.5-7B-Instruct responses, with four comparisons for each of 24 source items. Two members of the research team independently select Response A, Response B, or Tie, and disagreements are resolved by a third member of the research team. Because response order is randomized, automated preferences are mapped to the same displayed A/B orientation before comparison with the adjudicated human labels.

Closed-ended accuracy and open-ended evaluation scores are analyzed within their respective protocols. Open-ended scores are interpreted as comparative evaluation outcomes whose values also depend on the associated judge, constraint-checking, or preference-comparison procedure, rather than as exact measures of model capability. These design and annotation details are summarized in Table~\ref{tab:validation-overview}.
\begin{center}
\captionof{table}{Design and annotation coverage of the five supporting validation studies.}
\label{tab:validation-overview}

\includegraphics[width=\linewidth]{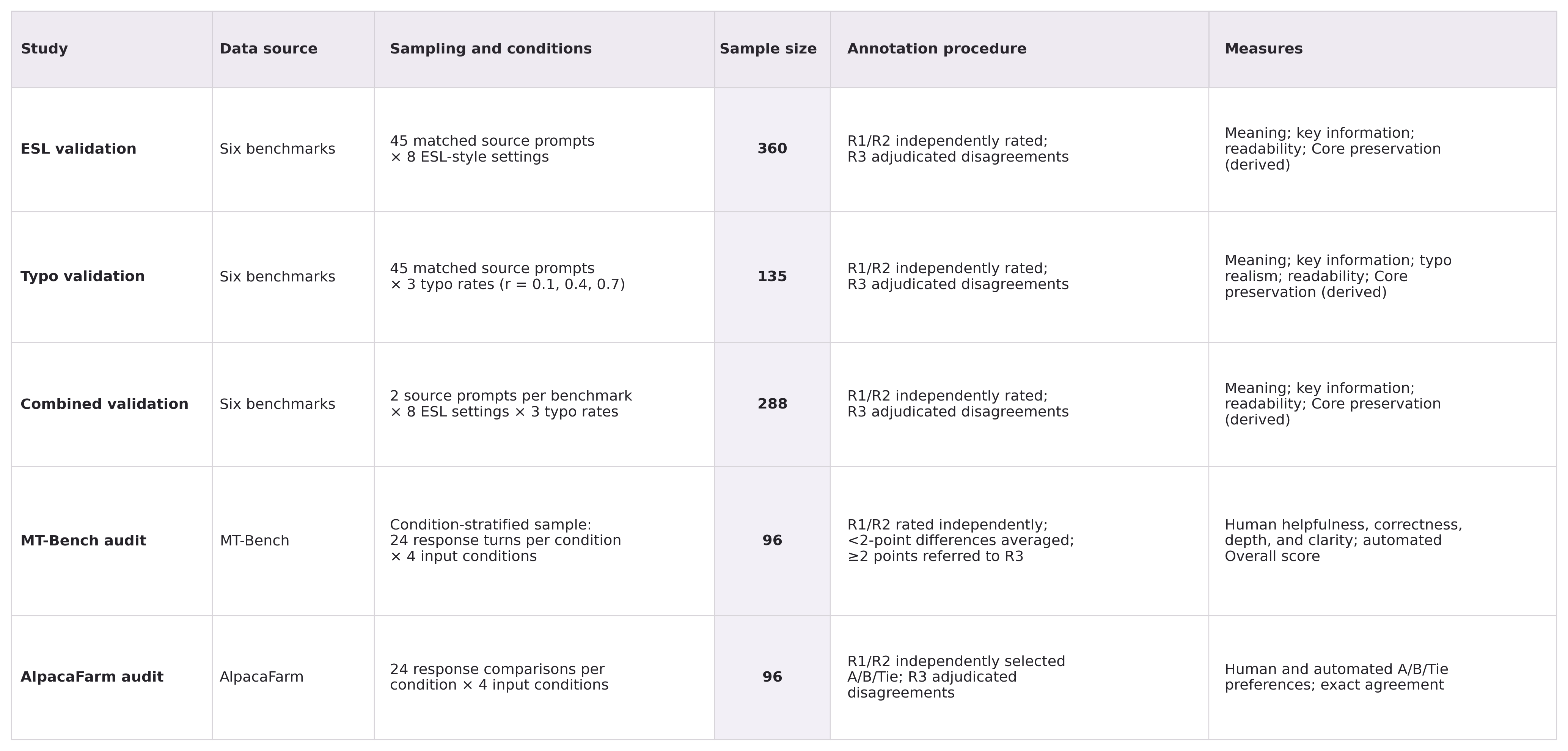}
\end{center}

\subsection{Degradation and interaction metrics}

To compare isolated and joint perturbation effects, we measure degradation relative to the clean Standard American English (SAE) baseline. Let \(P(\mathrm{SAE})\) denote model performance under the clean baseline, \(P(\mathrm{ESL})\) denote performance under the ESL-only condition, \(P(\mathrm{Typo},r)\) denote performance under typo-only input at corruption rate \(r\), and \(P(\mathrm{ESL}+\mathrm{Typo},r)\) denote performance under the combined condition.

We define the isolated and combined degradation terms as:
\[
\Delta_{\mathrm{ESL}} = P(\mathrm{SAE}) - P(\mathrm{ESL}),
\]
\[
\Delta_{\mathrm{Typo},r} = P(\mathrm{SAE}) - P(\mathrm{Typo},r),
\]
\[
\Delta_{\mathrm{Comb},r} = P(\mathrm{SAE}) - P(\mathrm{ESL}+\mathrm{Typo},r).
\]

We then define the interaction score as:
\[
\delta = \left(\Delta_{\mathrm{ESL}}+\Delta_{\mathrm{Typo},r}\right)-\Delta_{\mathrm{Comb},r}.
\]

Positive \(\delta\) indicates subadditivity, meaning that the combined degradation is smaller than the sum of the isolated degradations. Negative \(\delta\) indicates superadditivity, meaning that the combined degradation is larger than the sum of the isolated degradations. Values near zero indicate approximately additive behavior.

Because benchmark scores are bounded, we interpret \(\delta\) as a descriptive score-level measure of the relationship between isolated and combined perturbations.
\subsection{Statistical reporting}

All analyses are descriptive. We report means and standard deviations, with confidence intervals where applicable. Results are summarized within the relevant grouping variables, including model, benchmark, ESL-style variant, perturbation condition, and typo rate. All comparisons are computed within matched model--benchmark settings so that isolated and combined perturbation effects remain directly comparable.

\section{Results}

\begin{figure}[htbp]
\centering
\includegraphics[width=0.7\linewidth]{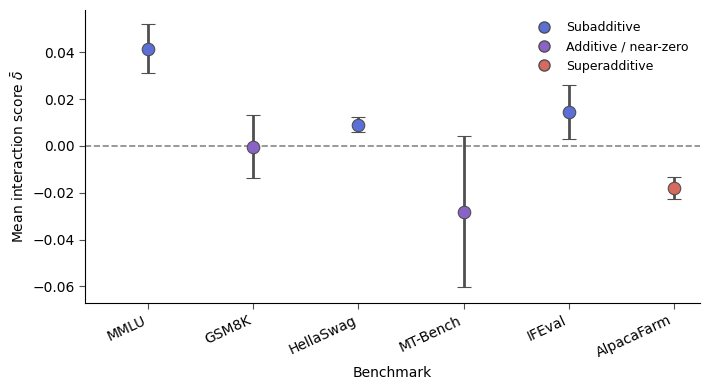}
\caption{Interaction patterns across benchmarks. Points show arithmetic mean interaction scores $\bar{\delta}$ computed from the corresponding model--ESL-style-setting--typo-rate cells, and error bars show one sample standard deviation across those cells. The dashed horizontal line marks $\delta=0$, corresponding to additive behavior on the benchmark-score scale.}
\label{fig:interaction_summary}
\end{figure}

The matched comparisons show that the combined condition generally produces a larger performance decrease than either isolated condition. However, the relationship between the combined degradation and the additive prediction varies across benchmarks. Figure~\ref{fig:interaction_summary} summarizes the benchmark-level interaction scores. MMLU shows a positive interaction, corresponding to subadditive degradation. HellaSwag also shows a positive deviation, whereas GSM8K remains close to additive. The open-ended benchmarks are less uniform: MT-Bench and AlpacaFarm have negative benchmark-level estimates, whereas IFEval has a positive estimate. Thus, the direction of the interaction and the within-benchmark variability of the estimates vary across tasks and evaluation procedures.

These results indicate that performance under either isolated transformation does not consistently characterize performance when both transformations are present. The human-validation analyses further qualify the interpretation of these estimates by showing that the measured effects reflect the complete transformation pipeline, including any associated change in task recoverability.

\subsection{Closed-ended tasks}

\subsubsection{MMLU: Factual knowledge}

\begin{figure}[!htbp]
\centering
\includegraphics[width=\linewidth]{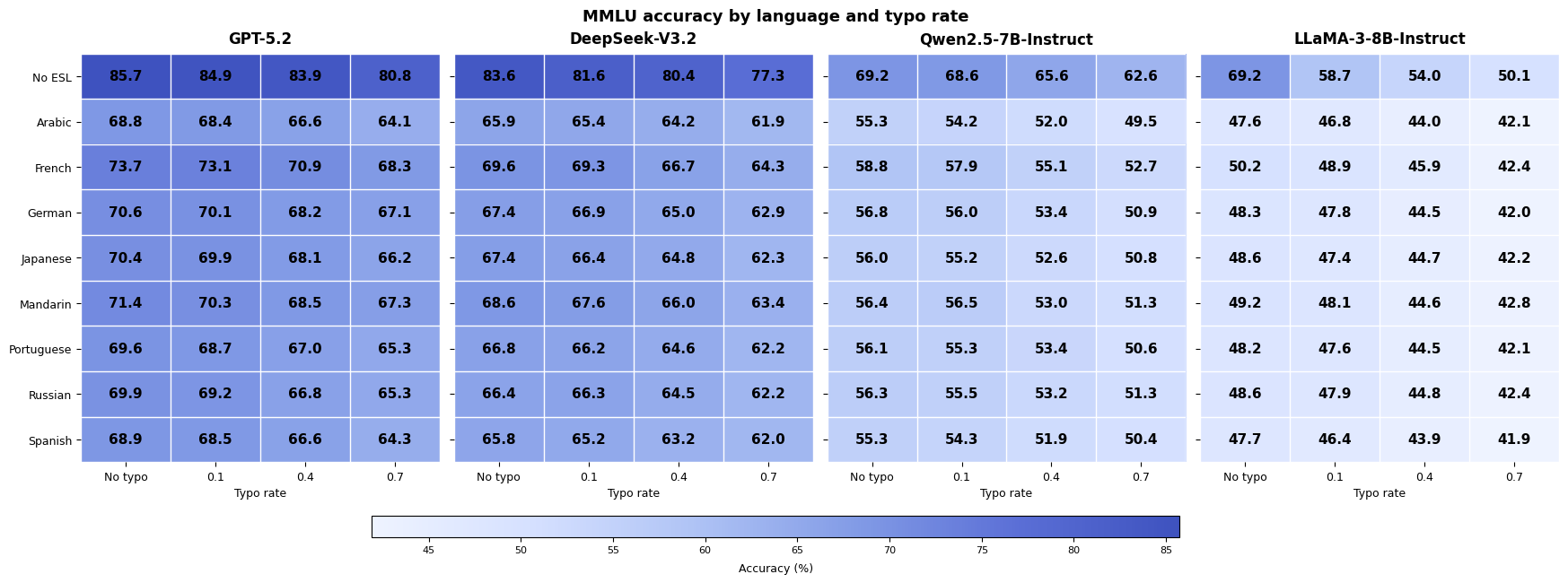}
\caption{MMLU accuracy percentages by transformation setting and typo rate for each model.}
\label{fig:figure2}
\end{figure}

We first examine MMLU, a broad multiple-choice knowledge benchmark. As shown in Figure~\ref{fig:figure2}, accuracy decreases relative to the clean benchmark-English baseline when either ESL-style variation or typographical perturbation is introduced, and the combined condition generally produces the lowest accuracy. Because the source question, answer options, and scoring rule are held fixed across conditions, these comparisons provide a direct description of model performance under the operationalized transformations.

The interaction estimates vary across models. Qwen2.5-7B-Instruct and DeepSeek-V3.2 remain close to additive, with mean interaction scores of approximately $\bar{\delta}=0.004\pm0.007$ and $\bar{\delta}=0.013\pm0.006$, respectively. LLaMA-3-8B-Instruct shows a positive mean interaction ($\bar{\delta}\approx 0.112 \pm 0.014$), corresponding to subadditive degradation. GPT-5.2 remains approximately additive on average ($\bar{\delta}\approx -0.00046 \pm 0.00662$).Under the sign convention defined in the Methods, a positive $\delta$ indicates that the combined degradation is smaller than the sum of the two isolated degradations.

The MMLU results also vary across the eight synthetic transformation settings. At $r=0.7$, the Arabic-informed, Spanish-informed, and Portuguese-informed settings have lower accuracies than the French-informed setting in several model panels. This pattern indicates heterogeneity across the experimental transformation settings; its broader interpretation is considered in the Discussion.

\subsubsection{GSM8K: Mathematical reasoning}

\begin{figure}[htbp]
\centering
\includegraphics[width=\linewidth]{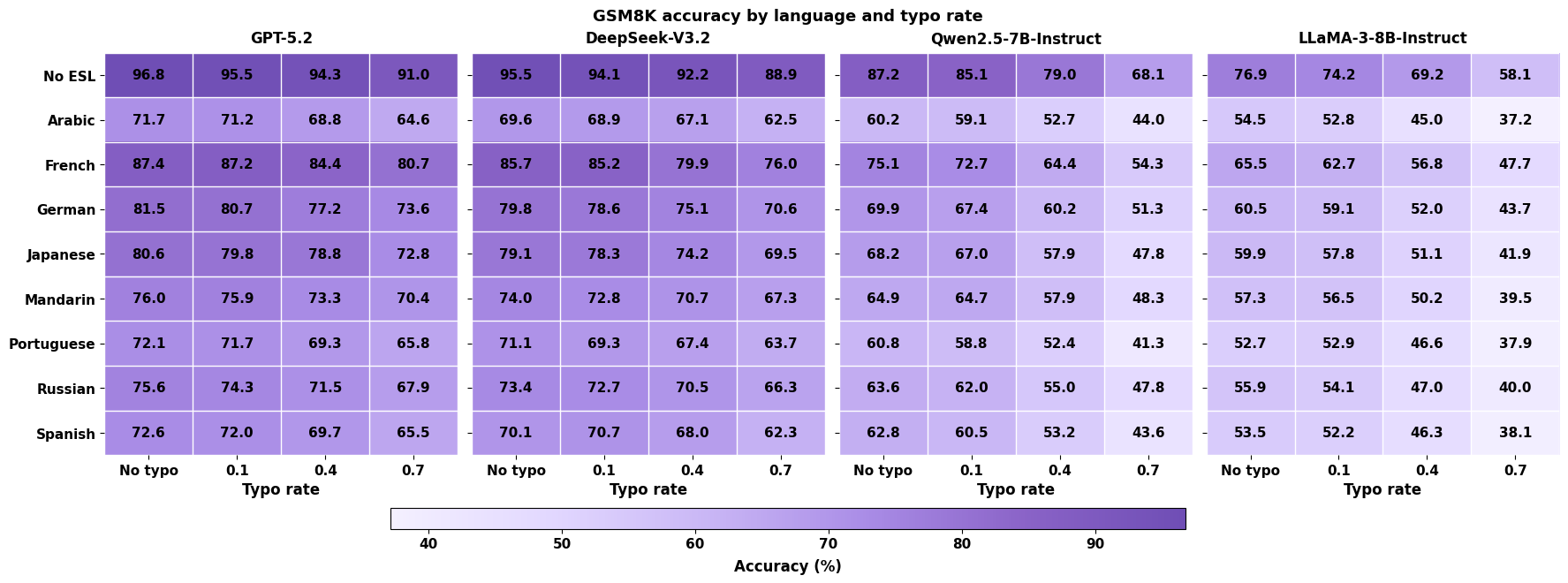}
\caption{GSM8K accuracy percentages by transformation setting and typo rate for each model.}
\label{fig:gsm8k}
\end{figure}

We next examine GSM8K, a benchmark of multi-step mathematical reasoning. Across the matched model--setting--typo-rate cells shown in Figure~\ref{fig:gsm8k}, the mean interaction is close to zero, with $\bar{\delta}\approx0.001$ and $\mathrm{SD}\approx0.057$. The combined degradation is therefore approximately additive on average.

At the severe stress-test rate $r=0.7$, the combined condition produces substantial accuracy loss relative to the clean benchmark-English baseline. Across the four model panels, the largest average drops occur for the Arabic-informed, Portuguese-informed, and Spanish-informed settings, at approximately $0.36$--$0.37$, followed by the Russian-informed setting at approximately $0.34$. The French-informed setting shows the smallest average drop.

\subsubsection{HellaSwag: Commonsense reasoning}

\begin{figure}[htbp]
\centering
\includegraphics[width=\linewidth]{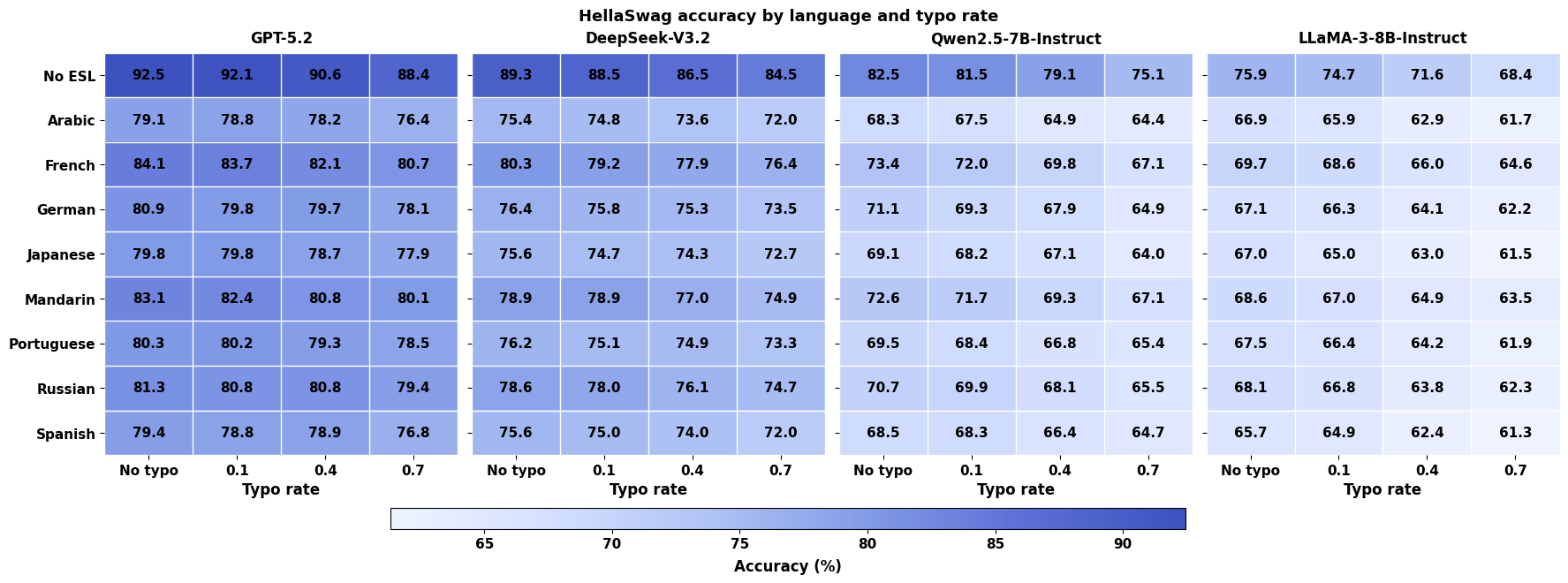}
\caption{HellaSwag accuracy percentages by transformation setting and typo rate for each model.}
\label{fig:figure4}
\end{figure}

We next examine HellaSwag, a commonsense reasoning benchmark in which the model selects the most plausible continuation. As shown in Figure~\ref{fig:figure4}, accuracy generally decreases from the clean condition to the isolated transformation conditions and then to the combined ESL-style and typographical condition. The absolute degradation is smaller than that observed for MMLU and GSM8K.

The benchmark-level interaction is subadditive, with $\bar{\delta}\approx0.009$ and $\mathrm{SD}\approx0.016$. This estimate is smaller than the positive interaction observed for MMLU but larger than the near-zero estimate for GSM8K.

At $r = 0.7$, the Arabic and Spanish ESL-style settings show the largest average combined drops, with the Japanese ESL-style setting close behind.

\subsection{Open-ended tasks}

\subsubsection{MT-Bench: Multi-turn conversation and generation}

MT-Bench assesses the quality of multi-turn conversational responses using an automated LLM judge. Responses are scored on helpfulness, correctness, depth, and clarity, together with an overall score used in the primary analysis.

\begin{center}
\captionof{table}{Average MT-Bench performance under the SAE baseline, ESL-only inputs, and typo-only inputs at $r=0.7$.}
\label{tab:mtbench_isolated}

\includegraphics[width=\linewidth]{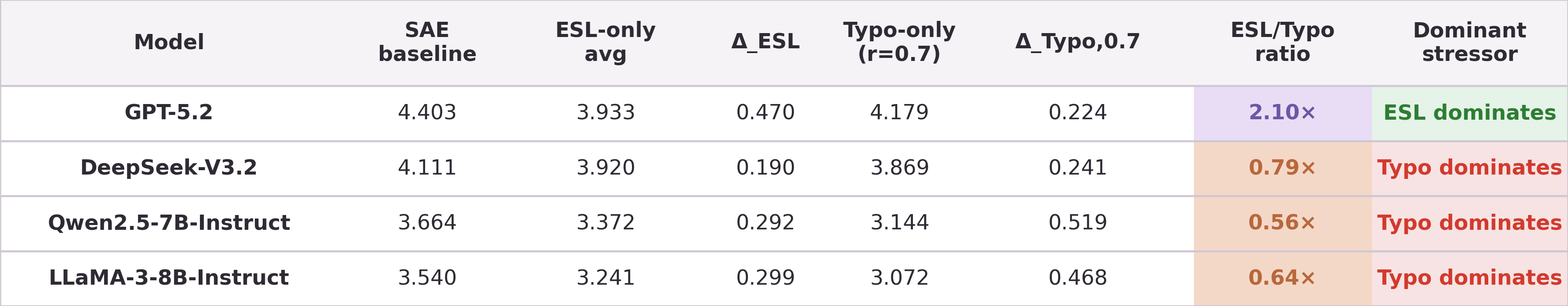}
\end{center}
As shown in Table~\ref{tab:mtbench_isolated}, both ESL-style variation and typographical perturbation reduce MT-Bench scores, although the relative magnitude of the two isolated degradations differs across models. The results therefore show model-specific differences in robustness to the two transformation types.

The interaction estimates also vary substantially across models. At the benchmark level, the arithmetic mean interaction is close to additive, with $\bar{\delta}\approx-0.028$ and $\mathrm{SD}\approx0.162$. Qwen2.5-7B-Instruct shows a positive mean interaction, with $\bar{\delta}\approx0.123$ and $\mathrm{SD}\approx0.171$, corresponding to subadditive degradation. LLaMA-3-8B-Instruct remains close to additive, with $\bar{\delta}\approx-0.028$ and $\mathrm{SD}\approx0.106$. DeepSeek-V3.2 and GPT-5.2 show negative mean interactions, with $\bar{\delta}\approx-0.149$ and $\mathrm{SD}\approx0.104$, and $\bar{\delta}\approx-0.0585$ and $\mathrm{SD}\approx0.129$, respectively, corresponding to superadditive degradation. The benchmark-level mean therefore masks substantial model-level heterogeneity.

\begin{figure}[htbp]
\centering
\includegraphics[width=0.4\linewidth]{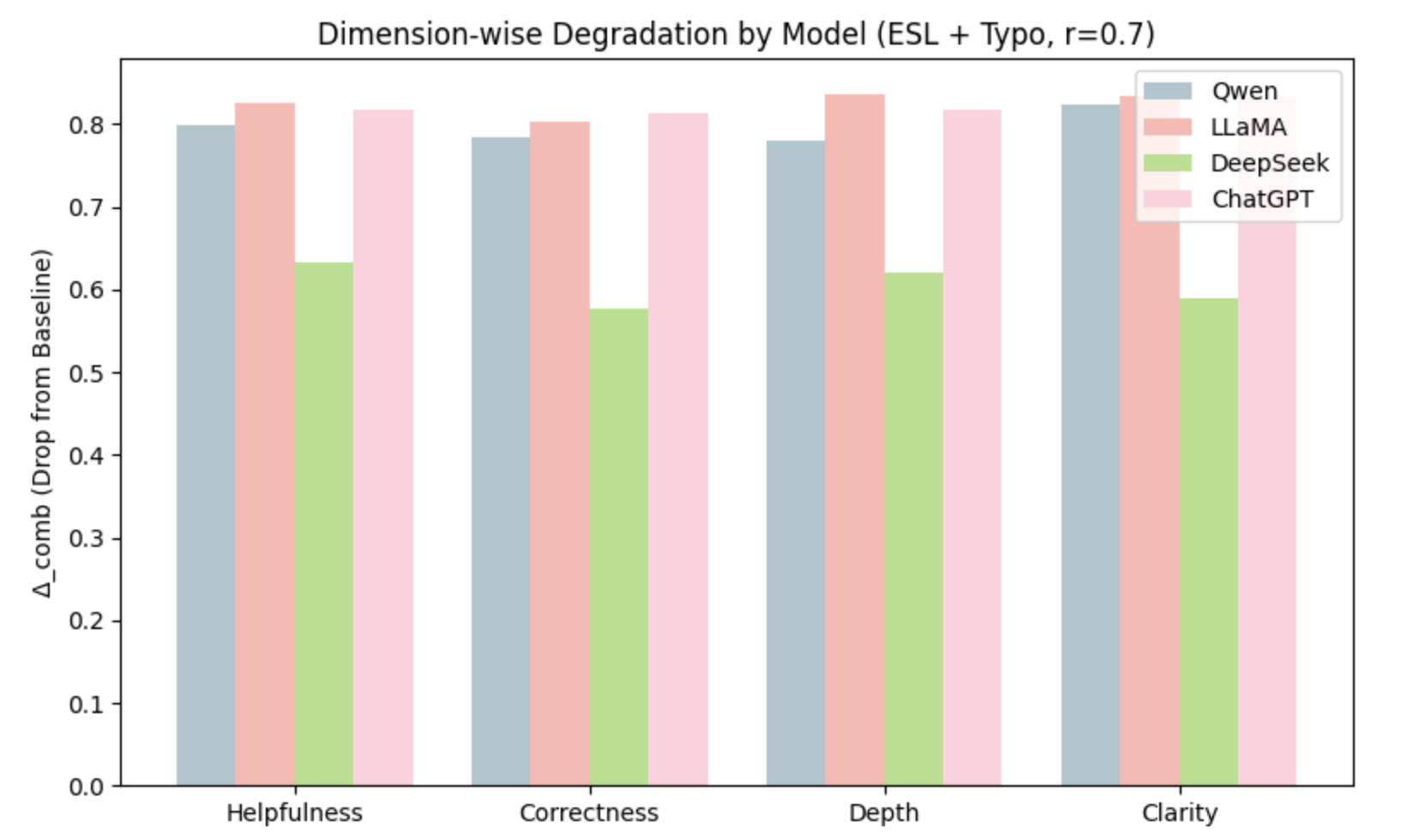}
\caption{Dimension-wise degradation on MT-Bench under combined ESL-style and typographical perturbations at $r=0.7$.}
\label{fig:mtbench_category}
\end{figure}

Figure~\ref{fig:mtbench_category} shows degradation across the four MT-Bench scoring dimensions under the combined condition. At $r=0.7$, mean degradation is similar across helpfulness, correctness, depth, and clarity. The decline in judged response quality is therefore distributed across the evaluation dimensions rather than concentrated in a single category. DeepSeek-V3.2 shows the smallest degradation across the plotted dimensions within this comparison.

Because MT-Bench relies on an automated judge, correspondence between automated and human scores is examined separately in Section~\ref{sec:judge_validation}.

\subsubsection{IFEval: Instruction following}

IFEval evaluates whether model responses satisfy explicit, verifiable instructions using deterministic rule-based checks. A response is counted as successful only when all associated constraints are satisfied.

\begin{tableorg}[htbp]
\centering
\includegraphics[width=\linewidth]{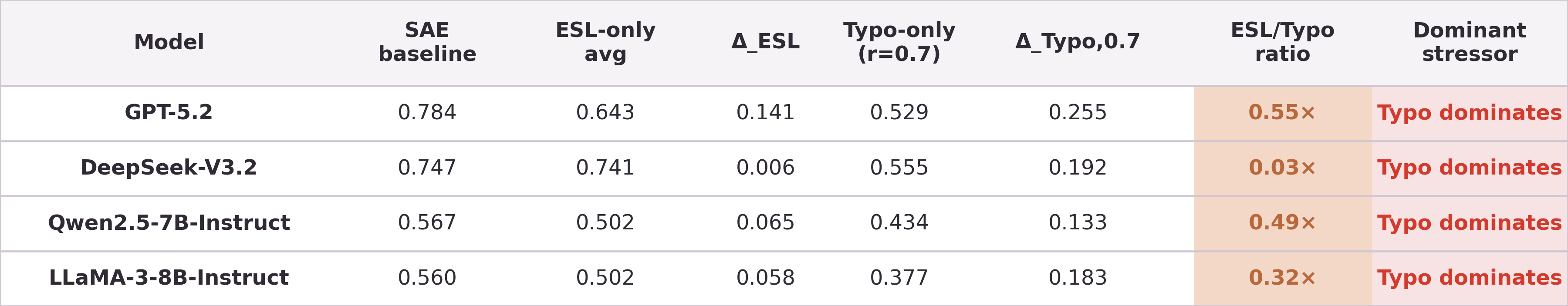}
\caption{IFEval performance under the clean, ESL-only, and typo-only conditions at $r=0.7$. ESL-only scores are averaged across the eight transformation settings.}
\label{tab:ifeval_isolated}
\end{tableorg}

At the severe stress-test rate $r=0.7$, typographical perturbation produces greater degradation than ESL-style variation for all four models. As shown in Table~\ref{tab:ifeval_isolated}, $\Delta_{\mathrm{Typo},0.7}$ ranges from approximately $0.13$ to $0.26$, whereas $\Delta_{\mathrm{ESL}}$ ranges from approximately $0.006$ to $0.141$. For GPT-5.2, typo-only degradation is approximately $0.255$, compared with approximately $0.141$ under ESL-only variation. The same ordering is observed for DeepSeek-V3.2, Qwen2.5-7B-Instruct, and LLaMA-3-8B-Instruct. All four models have ESL/Typo degradation ratios below 1 at $r=0.7$.

Across the three typo rates ($r \in \{0.1, 0.4, 0.7\}$), the mean interaction is close to zero, with $\bar{\delta} \approx 0.010$ and $\mathrm{SD} \approx 0.054$. The combined condition is therefore approximately additive on average, even though typo-only inputs produce the larger isolated performance decrease. This pattern is consistent with the structure of the IFEval evaluation protocol. Typographical perturbations may alter keywords or other surface cues associated with the explicit constraints checked by the evaluator. The human-validation results similarly show lower task-content preservation as typo severity increases.

\subsubsection{AlpacaFarm: Preference-based evaluation}

AlpacaFarm evaluates response quality through pairwise preference comparisons rather than exact correctness or deterministic instruction-following checks.

\begin{tableorg}[htbp]
\centering
\includegraphics[width=\linewidth]{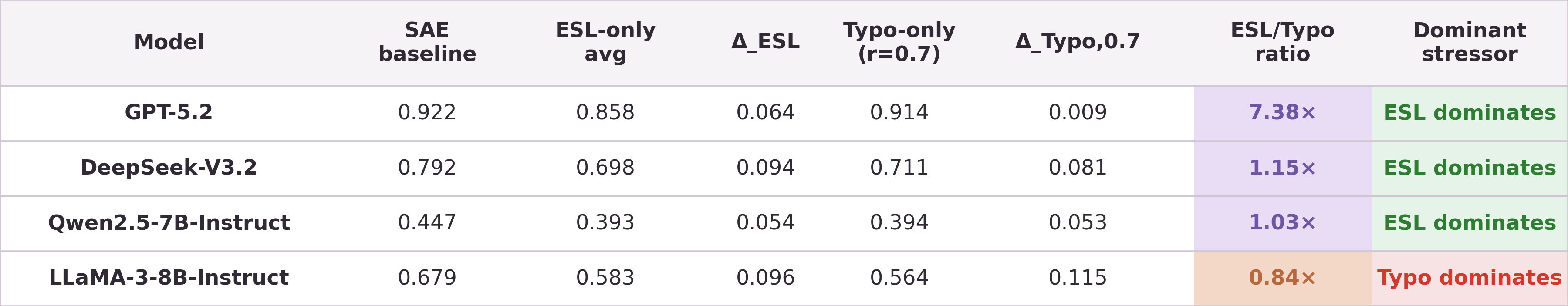}
\caption{AlpacaFarm performance under the clean, ESL-only, and typo-only conditions at $r=0.7$. ESL-only scores are averaged across the eight transformation settings.}
\label{tab:alpacafarm_isolated}
\end{tableorg}

As shown in Table~\ref{tab:alpacafarm_isolated}, at $r=0.7$, the isolated ESL-style degradations are approximately $0.064$, $0.094$, $0.054$, and $0.096$ for GPT-5.2, DeepSeek-V3.2, Qwen2.5-7B-Instruct, and LLaMA-3-8B-Instruct, respectively. The corresponding typo-only degradations are approximately $0.009$, $0.081$, $0.053$, and $0.115$. LLaMA-3-8B-Instruct shows the largest combined degradation, with $\Delta_{\mathrm{Comb},0.7}\approx0.244$.

The benchmark-level interaction is weakly superadditive, with $\bar{\delta}\approx-0.018$ and $\mathrm{SD}\approx0.022$. Model-level mean interactions are also negative but differ in magnitude. DeepSeek-V3.2 and LLaMA-3-8B-Instruct show relatively small negative estimates, with $\bar{\delta}\approx-0.008$ and $\bar{\delta}\approx-0.011$, respectively. Qwen2.5-7B-Instruct and GPT-5.2 show larger negative estimates, with $\bar{\delta}\approx-0.024$ and $\bar{\delta}\approx-0.027$.

The AlpacaFarm results therefore show a modest superadditive interaction under preference-based evaluation. Together with the MT-Bench and IFEval results, these findings show that the direction and magnitude of the interaction differ across open-ended evaluation protocols.

\subsection{Supporting validation analyses}
\label{sec:validation_results}

We conduct two supporting validation analyses. First, we assess task preservation for the ESL-style and typographical transformations separately by comparing each transformed prompt with its clean source prompt. Second, for MT-Bench and AlpacaFarm, we examine the correspondence between automated evaluations and human judgments. The prompt-level analyses assess the ESL-only, typo-only, and combined conditions in their respective validation samples. Their sampling and aggregation procedures are reported separately.

\subsubsection{Human validation of synthetic perturbations}
\label{sec:perturbation_validation}

\begin{figure}[!htbp]
    \centering
    \includegraphics[width=0.96\linewidth]{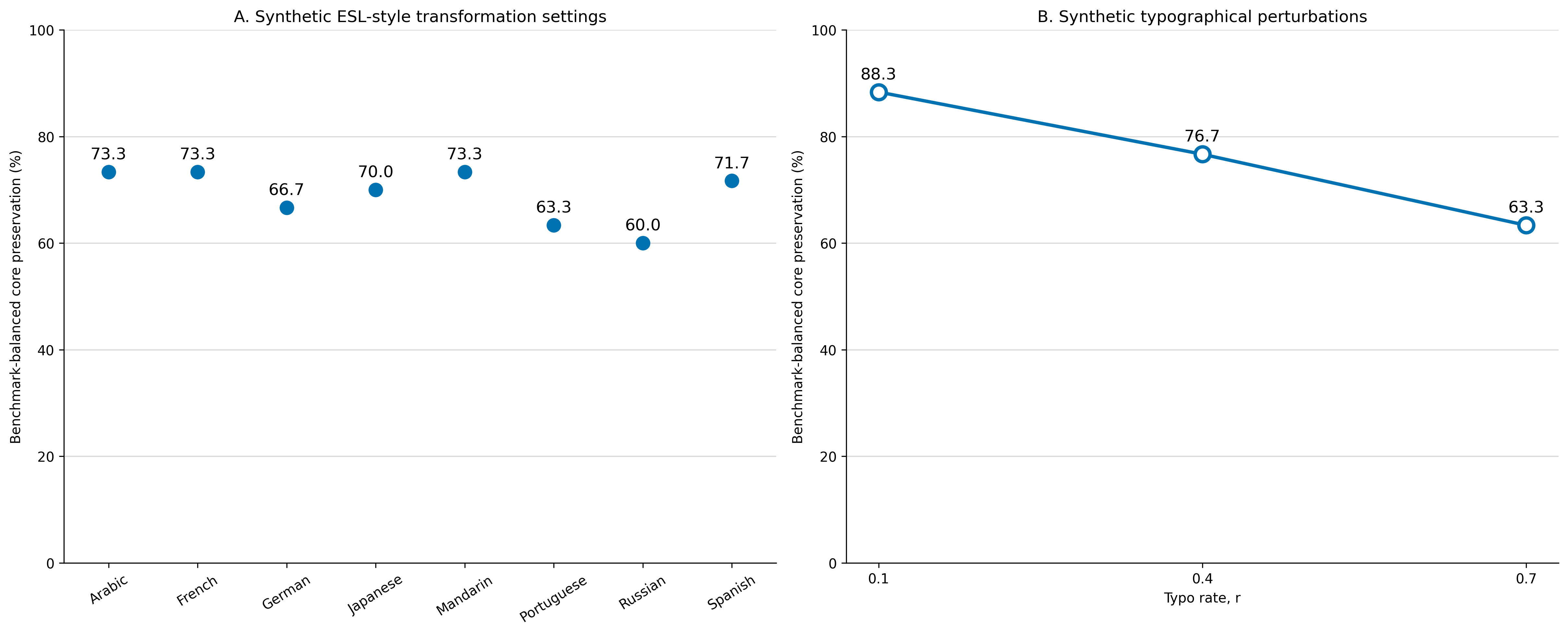}
    \caption{Human validation of synthetic input perturbations.
    \textbf{(A)} Benchmark-balanced macro-average core-content
    preservation across eight synthetic ESL-style transformation
    settings. \textbf{(B)} The same measure across typographical
    corruption rates $r \in \{0.1,0.4,0.7\}$. Core-content preservation
    requires both meaning preservation and key-information preservation.
    The $r=0.7$ condition is a severe synthetic stress test.}
    \label{fig:perturbation_validation}
\end{figure}

We first assess whether the synthetic transformations preserve the meaning and task-relevant key information of the original prompts. As shown in Figure~\ref{fig:perturbation_validation}, benchmark-balanced macro-averaged core-content preservation ranges from 60.0\% to 73.3\% across the eight synthetic ESL-style transformation settings. For typographical perturbations, preservation decreases with severity, from 88.3\% at $r=0.1$ to 76.7\% at $r=0.4$ and 63.3\% at $r=0.7$.

Core-content preservation requires preservation of both the overall meaning and the task-relevant key information. These findings qualify the interpretation of the benchmark results. The reported performance changes reflect the complete operationalized transformation conditions, including any associated reduction in task recoverability, rather than surface-form variation alone.

\begin{figure}[htbp]
\centering
\includegraphics[width=0.85\linewidth]{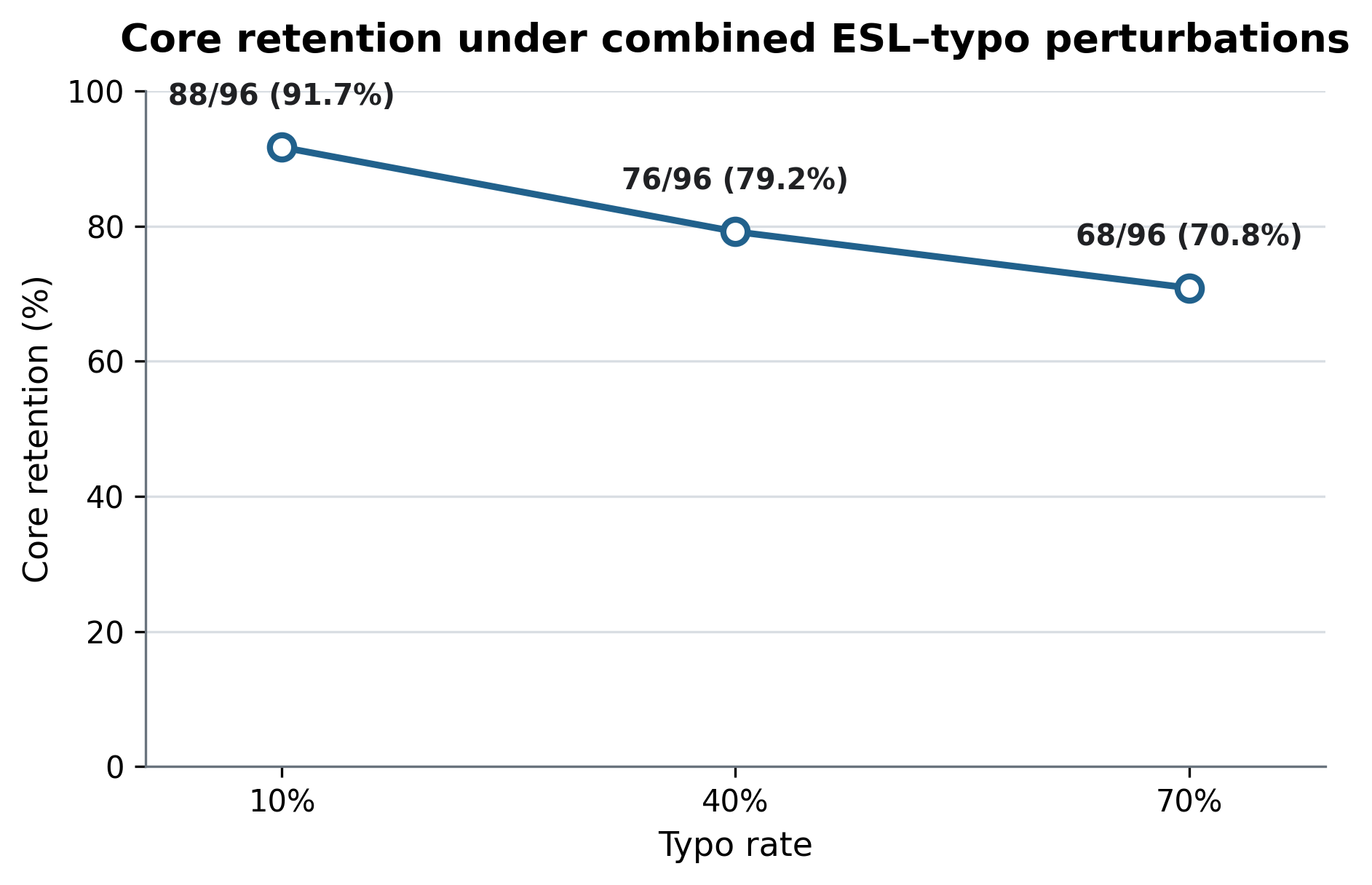}
\caption{Core-content preservation in the combined ESL-style and typo validation sample (96 transformed prompts per rate). A prompt was counted as preserved only when both meaning and key information were rated Yes. The $r=0.7$ condition is a severe synthetic stress test.}
\label{fig:combined-retention}
\end{figure}

We also examined core-content retention in the combined
ESL-style-plus-typo condition (Figure~\ref{fig:combined-retention}).
Of the 96 prompts assessed at each typo rate, both meaning and
key information were preserved in 88 (91.7\%) at $r=0.1$,
76 (79.2\%) at $r=0.4$, and 68 (70.8\%) at $r=0.7$.
Core-content retention therefore declined as typo severity
increased within this sample.The combined and ESL-only audits used different source-prompt samples, so their preservation rates cannot be directly compared to estimate the effect of adding typos.

\subsubsection{Human audit of automated evaluation}
\label{sec:judge_validation}

\begin{figure}[htbp]
\centering
\includegraphics[width=\linewidth]{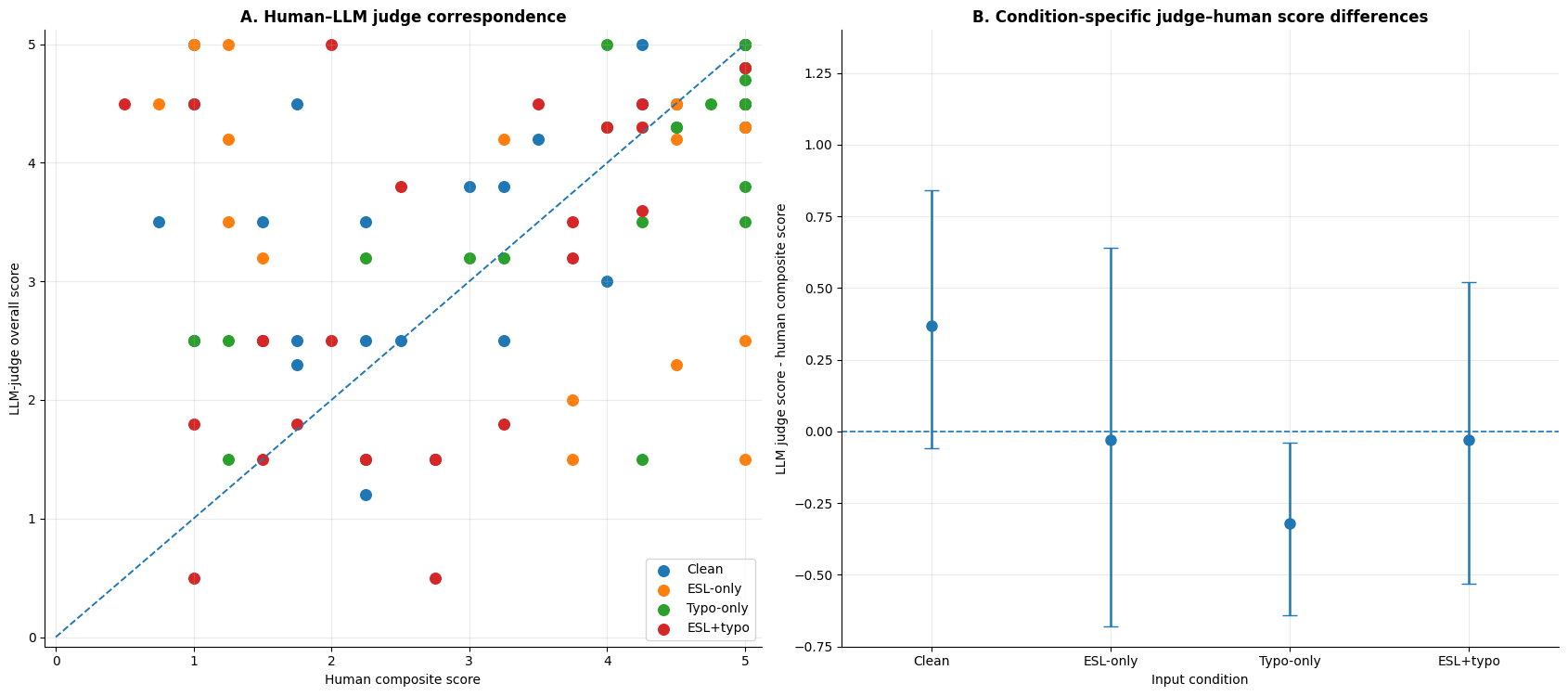}
\caption{Human audit of automated MT-Bench evaluation. (A) Final human composite scores versus LLM-judge overall scores; the dashed diagonal marks $y=x$. (B) Mean signed differences, defined as the LLM-judge score minus the human score. Error bars show within-condition percentile-based 95\% bootstrap confidence intervals from 10,000 resamples. Positive values indicate higher scores from the automated judge.}
\label{fig:judge_validation}
\end{figure}

We next compare automated evaluations with human judgments for MT-Bench and AlpacaFarm. For MT-Bench, the audit comprises 96 response-turns, with 24 observations from each input condition. Automated judge scores show a moderate rank association with final human composite scores, with Spearman's $\rho=0.44$ and a mean absolute error of 0.90 points on the 0--5 scale (Figure~\ref{fig:judge_validation}A).
Condition-specific mean judge–human differences and bootstrap intervals are reported in Table~\ref{tab:human-agreement}, Panel A. Because the audit was condition-stratified rather than organized into complete matched sets, these estimates are descriptive.

\FloatBarrier
For AlpacaFarm, automated and adjudicated human preference judgments agree in 69 of 96 pairwise comparisons (71.9\%; unweighted Cohen's $\kappa=0.53$). Figure~\ref{fig:alpaca-human-audit} shows the condition-specific exact agreement rates between automated and final human preference judgments. Table~\ref{tab:human-agreement}, Panel B, reports the full preference distributions. These condition-specific values are descriptive and should not be interpreted as definitive evidence of differences in evaluator reliability.
\begin{center}
\begin{minipage}{\linewidth}
\centering
\captionof{table}{Correspondence between final human and automated evaluation in condition-stratified audit samples.}
\label{tab:human-agreement}
\includegraphics[width=\linewidth]{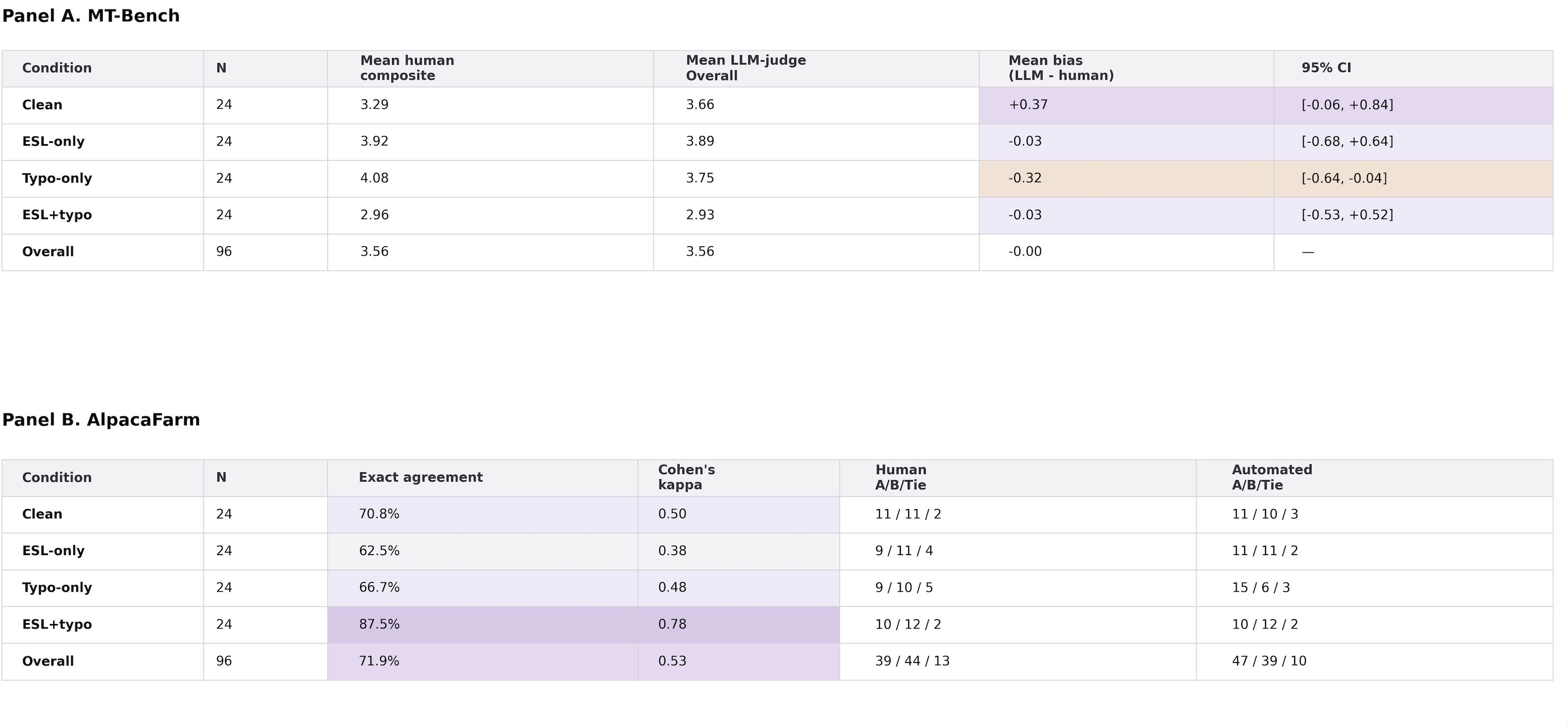}
\end{minipage}
\end{center}

\begin{center}
\begin{minipage}{\linewidth}
\centering
\includegraphics[width=\linewidth]{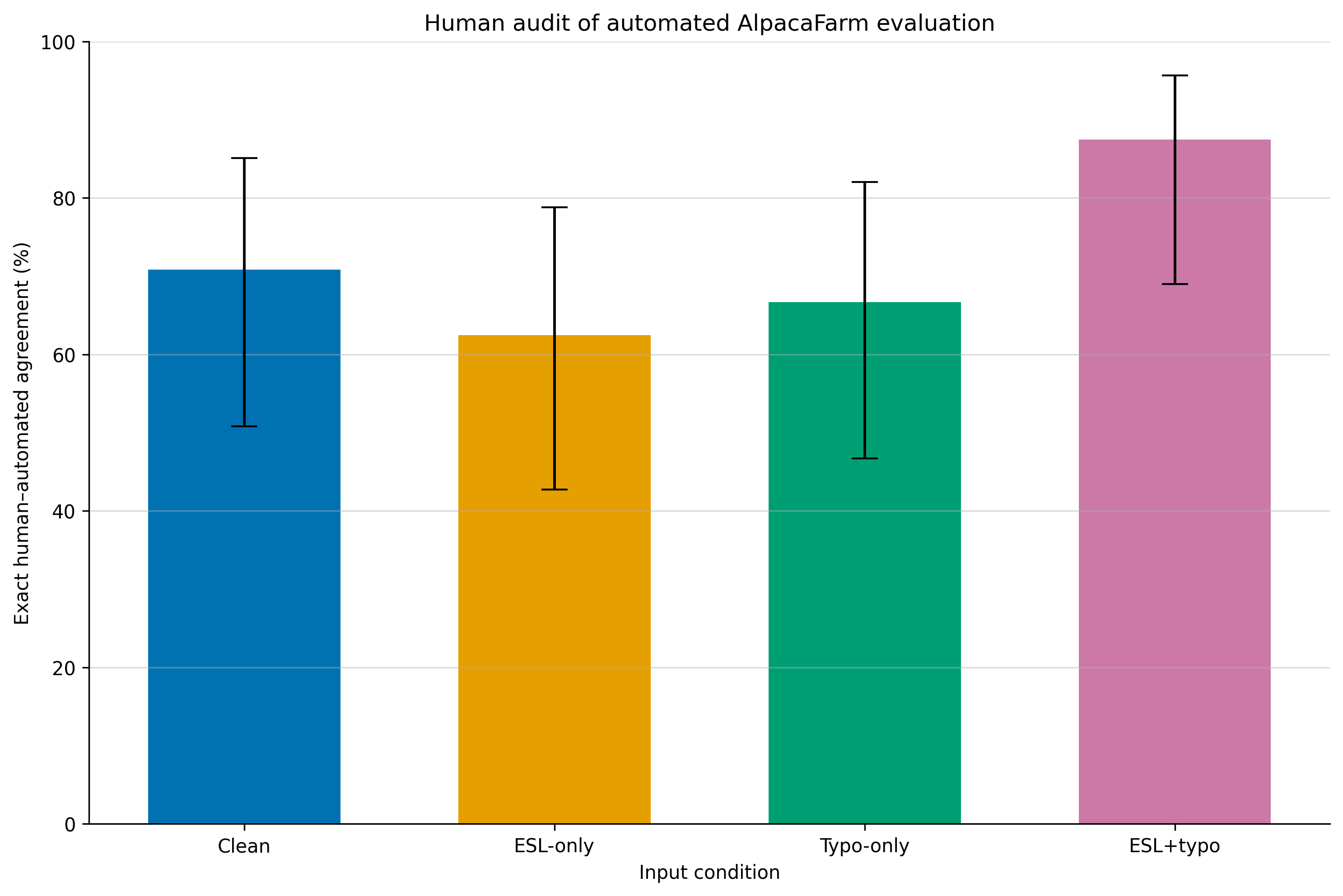}
\captionof{figure}{Human audit of automated AlpacaFarm evaluation. Bars show exact agreement between automated and final human A/B/Tie judgments within each input condition (\(n = 24\) per condition). Error bars denote 95\% Wilson confidence intervals. Overall, automated and human labels agreed on 69 of 96 comparisons (71.9\%; unweighted Cohen's \(\kappa = 0.53\); Table~\ref{tab:human-agreement}, Panel B).}
\label{fig:alpaca-human-audit}
\end{minipage}
\end{center}

The two audits show that automated evaluation is informative but does not perfectly correspond with human judgment. We therefore interpret MT-Bench and AlpacaFarm scores as protocol-dependent evaluation outcomes rather than as interchangeable substitutes for human assessment.

\FloatBarrier
\subsection{Cross-benchmark synthesis}

Across the closed-ended benchmarks, MMLU and HellaSwag show subadditive interactions, whereas GSM8K is approximately additive on average. The open-ended benchmarks show greater heterogeneity. MT-Bench exhibits substantial model-level variation, IFEval shows greater sensitivity to typo-only inputs while remaining close to additive at the aggregate level, and AlpacaFarm shows a weak superadditive interaction under preference-based evaluation. The direction and magnitude of the interaction therefore vary across benchmarks, models, and evaluation procedures.

Across the six benchmarks, performance under isolated ESL-style and typo-only conditions does not fully characterize performance when the two transformations are combined. Direct evaluation of the combined conditions therefore reveals robustness patterns that are not captured by either isolated transformation alone.

\section{Discussion}

\paragraph{Combined input variation changes the interpretation of robustness scores.}

The central finding is not simply that perturbed prompts reduce model performance. It is that the combined condition produces a pattern that cannot be recovered from either isolated condition alone. MMLU and HellaSwag showed subadditive interactions, GSM8K was approximately additive on average, and the open-ended benchmarks were more heterogeneous. These differences show that robustness is not a single context-free property of a model. In the present experiments, it is a property of the model, the task, the input condition, and the evaluation procedure considered together.

\paragraph{Benchmark format and evaluation protocol shape the observed pattern.}

The open-ended results make this point particularly clear. IFEval was more sensitive to typo-only than to ESL-only inputs, whereas MT-Bench showed model-specific differences in the relative effects of the two transformations. AlpacaFarm exhibited smaller and more balanced isolated changes, yet still showed weak superadditivity under preference-based evaluation. The same pair of input transformations therefore takes on different empirical meanings when success is defined by constraint satisfaction, judged response quality, or relative preference. Benchmark deltas consequently reflect the structure of the test as well as changes in model performance.

\paragraph{Implications for evaluation practice.}

When a reliability claim includes inputs in which multiple forms of variation may co-occur, the evaluation should examine relevant combinations or explicitly identify the resulting gap in coverage. Reporting matched clean, ESL-only, typo-only, and combined conditions allows the observed combined degradation to be compared with each isolated degradation and with their sum. Validation also belongs within this evaluation pipeline. The prompt-level analyses connect score changes to the preservation of task-relevant content, while the MT-Bench and AlpacaFarm audits place automated scores alongside independent human judgments. Together, these analyses make benchmark results more interpretable as evidence for specific evaluation conditions.

\paragraph{Evidential responsibility in LLM evaluation.}

Evaluators should make clear the conditions under which reliability claims are supported. They choose which inputs to test, how to construct transformations, and how to score responses, while downstream readers may see only aggregate results. Readers need this information to judge whether an evaluation is relevant to their intended use of a model. Evaluators therefore have an ethical responsibility to disclose limitations that could affect that judgment and to avoid giving a misleading impression of what the results show. Here, this means making clear that results from separate ESL-style and typo tests do not, by themselves, establish reliability when both forms of variation occur together. Claims about co-occurring variation therefore require tests of relevant combined conditions or an explicit statement that these conditions were not examined. If a transformation changes task-relevant content, that limitation should be reported rather than attributing every score change to linguistic form. Reports using automated judges should also explain how their scores correspond to human assessments and what remains uncertain. The aim is to define the scope of the evidence, not to test every possible input.

Building on work on transparent model reporting and measurement validity
\citep{mitchell2019model,jacobs2021measurement}, we distinguish three questions: whether the tested inputs cover the conditions named in a claim, whether transformations preserve the original task, and whether the scoring procedure supports the intended interpretation. Evidence for one question does not settle the others. Our combined-condition audit assesses task preservation only in sampled prompts; similarly, overall agreement between automated and human scores does not by itself establish that the evaluator supports comparisons across conditions.

These distinctions have practical consequences for reporting. Claims
covering co-occurring variation require evidence about relevant
combinations or an explicit statement that this coverage is missing.
Where task preservation is uncertain, score changes should be reported
as outcomes of the transformation pipeline rather than attributed
entirely to linguistic form. Where automated judgments are used,
their correspondence with human assessment and the limits of the
audit should accompany the results. The present study illustrates
these reporting obligations in a controlled setting; it does not
establish discrimination against ESL users or harm in deployment.
\paragraph{Limitations and future work.}

These findings concern a controlled testbed rather than the prevalence or lived context of naturally occurring user language. The transformations support matched comparisons, but they do not represent the full range of real ESL communication, code-switching, or everyday typing behavior; the \(r=0.7\) condition is deliberately a severe stress test. Prompt-level validation also included a sample of combined inputs at three typo rates. Its limited coverage does not establish task preservation for every evaluated prompt, benchmark, or ESL-style setting. The observed combined degradation may therefore reflect both sensitivity to input variation and changes in the recoverability of the original task. The interaction score describes performance under the generated conditions but does not separate these contributions. Bounded metrics, aggregation choices, and the use of automated or preference-based evaluation also shape the reported interaction patterns. Future work can extend this design to naturally occurring inputs collected with consent, broader model and task families, multilingual and code-switching settings, and additional forms of co-occurring input variation. Direct validation of combined inputs, followed by analyses restricted to task-preserving matched sets, would help assess whether the observed patterns persist when the original task remains recoverable.

\section*{Statements and Declarations}

\paragraph{Data availability}
The benchmark datasets used in this study (MMLU, GSM8K, HellaSwag, MT-Bench, IFEval, and AlpacaFarm) are available from their original sources. The processed perturbation sets, model outputs, and de-identified human-audit records supporting the findings of this study are available from the corresponding author upon reasonable request.

\paragraph{Code availability}
The code used to generate the input perturbations, run the model evaluations, compute the reported metrics, and produce the figures and tables will be made publicly available in a GitHub repository upon publication.


\bibliography{sn-bibliography}

\end{document}